# Ensemble of Convolutional Neural Networks Trained with Different Activation Functions

Gianluca Maguolo[a]*, Loris Nanni[a,] and Stefano Ghidoni[a]

[a] *University of Padua, Department of Information Engineering, viale Gradenigo 6, Padua, Italy.*

*Corresponding author

Email addresses:
Gianluca Maguolo: gianluca.maguolo@phd.unipd.it
Loris Nanni: loris.nanni@unipd.it
Stefano Ghidoni: stefano.ghidoni@unipd.it

ABSTRACT
Activation functions play a vital role in the training of Convolutional Neural Networks. For this reason, developing efficient and well-performing functions is a crucial problem in the deep learning community. The idea of these approaches is to allow a reliable parameter learning, avoiding vanishing gradient problems. The goal of this work is to propose an ensemble of Convolutional Neural Networks trained using several different activation functions. Moreover, a novel activation function is here proposed for the first time. Our aim is to improve the performance of Convolutional Neural Networks in small/medium sized biomedical datasets. Our results clearly show that the proposed ensemble outperforms Convolutional Neural Networks trained with a standard ReLU as activation function. The proposed ensemble outperforms with a p-value of 0.01 each tested stand-alone activation function; for reliable performance comparison we tested our approach on more than 10 datasets, using two well-known Convolutional Neural Networks: Vgg16 and ResNet50.
The MATLAB code used here will be available at https://github.com/LorisNanni.

## 1. Introduction and State of the Art

Neural networks are one of the most popular tools in artificial intelligence. In recent years they became the state of the art technique in many fields like image classification (He, Zhang, Ren, & Sun, 2016), object detection (Ren, He, Girshick, & Sun, 2015), face recognition (Schroff, Kalenichenko, & Philbin, 2015) and machine translation (Bahdanau, Cho, & Bengio, 2015). The first deep neural networks were trained using activation functions like the hyperbolic tangent or the sigmoid function. However, these functions saturate as the modulus of the input goes to infinity, while the gradients rapidly decrease, allowing only the training of shallow networks. In order to address these problems, in 2011 Glorot et al. (Glorot, Bordes, & Bengio, 2011) showed that deep networks can be efficiently trained using Rectified Linear Units (ReLU), an activation function which coincides with the identity function if the input is positive and it is zero when the input is negative (Nair & Hinton, 2010). Although this function is not differentiable, it outperformed the previous saturating activation functions, allowing AlexNet to win the ImageNet competition in 2012 (Krizhevsky, Sutskever, & Hinton, 2012). Since ReLU was very effective, very simple and very fast to evaluate, in the following years, many deep learning researchers focused on finding ReLU-like activations with slightly different properties.

One example is Leaky ReLU (Maas, 2013), an activation function that is equal to ReLU for positive inputs (i.e.: the identity function) and it has a very small slope $\alpha > 0$ for negative inputs, $\alpha$ being a hyperparameter. In this way, the gradient of the function is never zero and it is less likely that the optimization process gets stuck in local minima. The same idea is the basis for Exponential Linear Units (ELU) (Clevert, Unterthiner, & Hochreiter, 2015). ELU is once again equal to ReLU for positive inputs, but it exponentially decreases to a limit point $\alpha$ as the input goes to minus infinity. This means that this activation has always positive gradient, but, unlike Leaky ReLU, saturates on its left side. A famous modification of ELU is Scaled Exponential Linear Unit (SELU) (Klambauer, Unterthiner, Mayr, & Hochreiter, 2017), which is ELU multiplied by a constant $\lambda$. Their idea is to tune these hyperparameters in order to make SELU preserve the mean and the variance of its input features. This helps to mitigate the vanishing gradient problem and allows the authors to successfully train deep feed-forward networks.

Standard activation functions do not depend on any learnable parameters and the training of the network only modifies the weights and the biases. In 2015 He et al. (He, Zhang, Ren, & Sun, 2015) implemented

Parametric ReLU (PReLU), which is a Leaky ReLU activation function where the slope of the negative part is a learnable parameter. According to the authors, this idea was the key to reach super-human results in the ImageNet 2012 dataset. Since this method adds parameters to the network, this activation makes overfitting more likely, so it is suitable in particular for larger datasets. According to the authors, PReLU always outperforms non-learnable activations on the training set, but might fail to generalize on the test set. After that, many learnable activations with different shapes have been proposed (Agostinelli, Hoffman, Sadowski, & Baldi, 2014; Scardapane, Vaerenbergh, & Uncini, 2018). In particular, Agostinelli et al. (Agostinelli et al., 2014) proposed a piecewise linear activation that they called Adaptive Piecewise Linear Unit (APLU), whose slopes and points of non differentiability are learnt at training time. This method is the most similar to the one that we propose here.

Learnable activations can also be defined using multiple fixed activations as their starting point. Manessi and Rozza (Manessi & Rozza, 2018) created a new learnable activation function by learning an affine combination of tanh, ReLU and the identity function. More recently, it was proposed (Ramachandran, Zoph, & Le, 2017) the Swish activation function $f(x) = x\sigma(\beta x)$ where $\sigma(\cdot)$ is the sigmoid activation and $\beta$ is a parameter that can optionally be learnable. The authors found this activation function using reinforcement learning. They created a network that tried to create different activation functions with a reward related to the performance of the activation function chosen. They used very simple activations as building blocks that their network could use to generate more complex activations. According to them, the best performing function was the Swish activation. In the reinforcement learning framework, only standard activations were considered. This means that the Swish activation was found keeping the parameter $\beta$ fixed. However, in their tests the activation performed better if it was set to be learnable.

In this paper we propose a piecewise linear activation function which is the sum of PReLU and multiple Mexican hat functions: we named our approach Mexican ReLU (MeLU). It has a number of learnable parameters that ranges from zero to infinity. In our case, the total number of parameters is a hyperparameter. It is built to have desirable properties that can improve the representation power of the network and help the network to reach better minima. First of all, if the number of parameters goes to infinity, it can approximate every continuous function $f$ on a compact set. Moreover, it does not saturate in any direction and its gradient is almost never flat. Finally, modifying a parameter changes the activation only on a small interval, making the optimization process simpler. We introduce MeLU because the recent literature dealing with learnable activations showed that they have the potential to outperform standard activations, but their improvement over the state of the art did not lead to the replacement of ReLU as the standard activation in the deep learning community. We believe that it is because the overall improvement of the recently proposed activations over the state of the art is not that large to justify the use of more complex functions that have not been widely tested. With MeLU, we try to make the transition from ReLU to learnable activations easier by initializing MeLU so that it coincides with ReLU in the first training step, enabling efficient transfer learning. Besides, we show that ReLU and MeLU can be used together to create an ensemble.

The most important results of this work are the following:

- We compare several activation functions, using two different Convolutional Neural Networks (CNNs), in thirteen small/medium size biomedical dataset. The CNNs chosen for our tests are Vgg16 (Simonyan & Zisserman, 2015) and ResNet50 (He et al., 2016).
- We show that an ensemble of activation functions (AF) strongly outperforms each AF singularly considered.
- We propose a new activation function.

The rest of the paper is organized as follows. In Section 2 we describe the most popular activation functions in the literature. In Section 3 we introduce MeLU and present its most important properties. In Section 4, we evaluate our activation function in many different dataset classification tasks, and we compare it with other methods presented in Section 2. Conclusions and take-home messages are summarized in Section 5.

## 2. Activation functions for CNNs

In this section, we present some of the best performing activation functions proposed in the literature for deep neural networks. We compared these functions by substituting them into two well-known CNNs, ResNet50 and VGG16, pre-trained on ImageNet.

ResNet50 is a CNN whose main features are called skip connections (He et al., 2016). The difference with the usual building block of a standard CNN, namely a convolution followed by an activation, is that in a skip connection the input of a block is summed to its output. This should help the gradient flow.

VGG16 is a CNN whose blocks are made of small stacked convolutional filters (Simonyan & Zisserman, 2015). It has been shown that they have the same effect of larger convolutional filters, but they use less parameters.

## 2.1. Rectified Linear Units

Rectified Linear Unit (ReLU) is defined as

$$y_i = f(x_i) = \begin{cases} 0, & x_i < 0 \\ x_i, & x_i \geq 0 \end{cases} \quad \text{with} \quad f'(x_i) = \begin{cases} 0, & x_i < 0 \\ 1, & x_i \geq 0 \end{cases}. \tag{1}$$

## 2.2. Leaky ReLU

Leaky ReLU is defined as

$$y_i = f(x_i) = \begin{cases} ax_i, & x_i < 0 \\ x_i, & x_i \geq 0 \end{cases} \quad \text{with} \quad f'(x_i) = \begin{cases} a, & x_i < 0 \\ 1, & x_i \geq 0 \end{cases}, \tag{2}$$

where $a$ is a small real number. With respect to ReLU, this function has the advantage that there is no point where the gradient is null, helping the optimization process.

## 2.3. ELU

Exponential Linear Unit (ELU) is defined as

$$y_i = f(x_i) = \begin{cases} a(\exp(x_i) - 1), & x_i < 0 \\ x_i, & x_i \geq 0 \end{cases} \quad \text{with} \quad f'(x_i) = \begin{cases} a \exp(x_i), & x_i < 0 \\ 1, & x_i \geq 0 \end{cases}, \tag{3}$$

where $a$ is a real number. Like Leaky ReLU the gradient of this function is always positive. Besides it has the advantage of being differentiable. It also has the property of being bounded from below by $-a$.

## 2.4. SELU

Scaled Exponential Linear Unit (SELU) is defined as

$$y_i = f(x_i) = \begin{cases} sa(\exp x_i - 1), & x_i < 0 \\ sx_i, & x_i \geq 0 \end{cases} \quad \text{with} \quad \frac{dy_i}{dx_i} = f'(x_i) = \begin{cases} sa \exp(x_i), & x_i < 0 \\ s, & x_i \geq 0 \end{cases}, \tag{4}$$

where $a, s$ are real numbers. This function is basically ELU multiplied by an additional parameter. It was created in the context of feed-forward networks to avoid the problem of gradient vanishing or explosion. Klambauer sets the parameters $a = 1.6733$ and $s = 1.0507$ because this choice of the parameters allows SELU to map a random variable of null mean and unit variance in a random variable with null mean and unit variance.

## 2.5. PReLU

Parametric ReLU is defined as

$$y_i = f(x_i) = \begin{cases} a_c x_i, & x_i < 0 \\ x_i, & x_i \geq 0 \end{cases} \quad \text{with} \quad \frac{dy_i}{dx_i} = \begin{cases} a_c, & x_i < 0 \\ 1, & x_i \geq 0 \end{cases} \quad \text{and} \quad \frac{dy_i}{da_c} = \begin{cases} x_i, & x_i < 0 \\ 0, & x_i \geq 0 \end{cases}, \tag{5}$$

where $a_c$ are real numbers that are different for every channel of the input. The big difference between this function and Leaky ReLU is that the parameters $a_c$ are learnable.

## 2.6. S-Shaped ReLU (SReLU)

S-Shaped ReLU was firstly introduced in (Jin et al., 2016). It is defined as

$$y_i = f(x_i) = \begin{cases} t^l + a^l(x_i - t^l), & x_i < t^l \\ x_i, & t^l \leq x_i \leq t^r, \\ t^r + a^r(x_i - t^r), & x_i > t^r \end{cases} \quad (6)$$

where $t^l, t^r, a^l, a^r$ are learnable real numbers. This function has a very large representation power thanks to the high number of parameters. Its gradient is

$$\frac{dy_i}{dx_i} = f'(x_i) = \begin{cases} a^l, & x_i < t^l \\ 1, & t^l \leq x_i \leq t^r \\ a^r, & x_i > t^r \end{cases} \quad (7)$$

$$\frac{dy_i}{da^l} = \begin{cases} x_i - t^l, & x_i < t^l \\ 0, & x_i \geq t^l \end{cases} \quad (8)$$

$$\frac{dy_i}{dt^l} = \begin{cases} -a^l, & x_i < t^l \\ 0, & x_i \geq t^l \end{cases} \quad (9)$$

## 2.7. APLU

Adaptive Piecewise Linear Unit (APLU) is defined as

$$y_i = \text{ReLU}(x_i) + \sum_{c=1}^{n} a_c \max(0, -x_i + b_c), \quad (10)$$

where $a_c, b_c$ are real numbers that are different for every channel of the input. This function is piecewise linear and it can approximate any continuous function on a compact set, for a suitable choice of the parameters, as $n$ goes to infinity. The gradient of APLU is given by the sum of the gradients of ReLU and of the functions contained in the sum. The gradients of APLU with respect to the parameters are

$$\frac{df(x,a)}{da_c} = \begin{cases} -x + b_c, & x < b_c \\ 0, & x \geq b_c \end{cases}, \quad (11)$$

$$\frac{df(x,a)}{db_c} = \begin{cases} -a_c, & x < b_c \\ 0, & x \geq b_c \end{cases}, \quad (12)$$

Besides, a 0.001 $L^2$-penalty on the norm of the parameters $a_c$ is suggested in the original paper to avoid the explosion of the parameters. This means that there is an additional term in the loss function which is

$$L^{reg} = \sum_{c=1}^{n} |a_c|^2.$$

## 3. Mexican ReLU

In the previous sections, we introduced many activation functions that have already been proposed in the literature. In this section we introduce Mexican ReLU (MeLU), a new activation function for neural networks. MeLU relies on the same ideas that motivated the introduction of APLU, but it solves the problem of its unstable training without any penalty on the learnable parameters. Besides, in this paper we show that using different activation functions can be a simple and effective way to create an ensemble, hence a new well-performing activation could also be seen as a useful ingredient for building that ensemble. In order to define MeLU, let

$$\phi_{a,\lambda}(x) = \max(\lambda - |x - a|, 0) \tag{14}$$

be a "Mexican hat type" function, where $a, \lambda$ are real numbers. The name comes from the fact that this function is null when $|x - a| > \lambda$ and it constantly increases with a derivative of 1 between $a - \lambda$ and $a$ and decreases with a derivative of minus 1 between $a$ and $a + \lambda$. If one draws it, it has the shape of a Mexican hat. We are aware that the term Mexican hat refers to a famous wavelet in the field of computer vision, we chose to call $\phi_{a,\lambda}(x)$ "Mexican hat type" because its shape is similar to the shape of the wavelet. These functions are the building blocks of MeLU. MeLU is defined as

$$y_i = MeLU(x_i) = PReLU(x_i) + \sum_{j=1}^{k-1} c_j \, \phi_{a_j,\lambda_j}(x_i) \tag{15}$$

for each channel of the hidden layer. $k$ is total number of learnable parameters in every channel, i.e: the $k - 1$ coefficients of the Mexican hat functions plus one learnable parameter in PReLU. The parameters $c_j$ are learnable, $a_j, \lambda_j$ are fixed and they are chosen recursively.

At this point we must preliminarily introduce the parameter *maxInput*. *maxInput* is a parameter that depends on the upper bound of the inputs. For example, in case of RGB images, it should be to 255, while in case of audio signal it should be set to 1. We created MeLU depending on this parameter because the sum with the Mexican hat functions modify the original PReLU on a limited set and we use *maxInput* to decide how wide that set should be. Although we suggest to initialize *maxInput* depending on the upper bound on the inputs and although its name suggests to do that, *maxInput* can be initialized to any value. However, our experiments show that it works better when it is used as it was meant to be used.

The first Mexican hat function has its maximum in $2 \cdot maxInput$ and it is equal to zero in 0 and $4 \cdot maxInput$. The next two functions are chosen to be zero outside, respectively, the intervals $[0, 2 \cdot maxInput]$ and $[2 \cdot maxInput, 4 \cdot maxInput]$, and imposing that they have their maximum in $maxInput$ and $3 \cdot maxInput$. The next four Mexican hat functions are defined by iteratively dividing each interval into two parts. In Table 1 we summarize the value of the first seven Mexican hat functions.

Table 1. Fixed parameters of MeLU with $maxInput = 256$.

| j | 1 | 2 | 3 | 4 | 5 | 6 | 7 |
|---|---|---|---|---|---|---|---|
| $a_j$ | 512 | 256 | 768 | 128 | 384 | 640 | 896 |
| $\lambda_j$ | 512 | 256 | 256 | 128 | 128 | 128 | 128 |

We now show some properties of MeLU. The Mexican hat functions are continuous and piecewise differentiable, hence MeLU inherits these properties. If all the $c_i$ are initialized at zero, MeLU coincides with ReLU. This helps transfer learning when we substitute MeLU in a network pretrained with ReLU. The same holds for networks trained with Leaky ReLU or PReLU. Moreover, the derivatives of a Mexican hat functions are a Hilbert basis on a compact set with the $L^2$ norm, hence they can approximate every function in $L^2([0,1024])$ as $k$ goes to infinity.

It is worth noting that the structure of a hidden layer is $f(Ah + b)$, where $h$ is the input of the hidden layer, $A$ is the weight matrix, $b$ is the bias and $f$ is the activation. If we consider the joint optimization of the weights, the bias and the activation parameters, we see that we can approximate any continuous function on a compact set. Consider an interval I and let $g$ be a function on that interval. We can approximate $g$ using MeLU by simply choosing $A$, $b$ to map $I$ into $[0, 4 \cdot maxInput]$ falling in the previous case, which is the approximation on a compact set.

Let us now focus on the relationship between MeLU and the other activations in the literature. It is clear that MeLU extends ReLU, Leaky ReLU and PReLU. Since it has more parameters, it has a higher representation power, but it might overfit easily. The most similar activations to MeLU in the literature are S-shaped ReLU and APLU. S-shaped ReLU is somehow the dual of MeLU: S-shaped ReLU divides the real line into two half lines and an interval and changes the slope of the activation on the two half lines. Conversely, most of the basis functions of MeLU have a limited support.

APLU and MeLU look very similar. Indeed, they can approximate the same set of functions: piecewise linear functions which are equal to the identity for $x$ large enough. However, they do it in a very different way. For the right choice of the parameters, APLU can be equal to any piecewise linear function because the points of non-differentiability are learnable, while MeLU can represent every piecewise linear function only exploiting the joint optimization of the weights matrix and the biases. This means that MeLU adds to a network one half of the parameters of APLU and has the same representation power. The second difference between the two functions is in the gradients. In optimizing a neural network there are two important factors: the output of every hidden layer and the gradient of that output with respect to the parameters. The gradients of MeLU with respect to the parameters are the Mexican hat functions. The gradients of APLU are computed in Section 2.7. It is clear that they are very different. The strong point of the solution we propose is its superior performance at the optimization stage. Without loss of generality, suppose that at a certain point in the training process $b_1 > b_2 > \cdots > b_n$. Suppose now that it would be optimal for the network to modify the activation function between $b_1$ and $b_2$. The only parameters whose gradients are not null in that interval are $a_1$ and $b_1$, so changing them would be optimal. However, changing $a_1$ has a very small effect near $b_1$, since the modulus of its gradient is the distance between $x$ and $b_1$. The only option left would be changing $b_1$. However, even this option is not very good, since this would change the function even where it is not needed, since the support of the gradient is unbounded. This means that the optimization process might be very inefficient. Conversely, in the same situation, MeLU changes the activation exactly where it is needed, making the optimization easier.

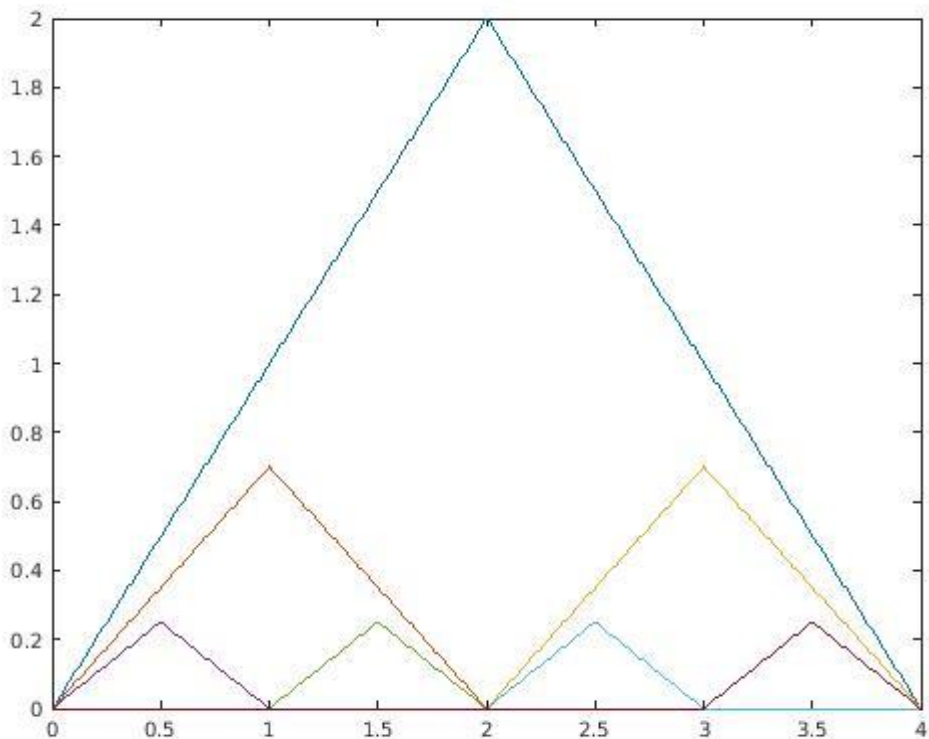

Figure 1. The first seven Mexican hat functions for the construction of MeLU (multiplied by a coefficient to avoid the overlap of the graphs). The input of the activation is on the horizontal axis, the output is on the vertical axis.

In the figure above, we can see the difference between the basis functions of MeLU and APLU. This might be the reason why the coefficients in APLU must be regularized with an $L^2$ penalty and benefit from a low learning rate, while MeLU does not need any regularization.

In our experiments we set $k = 4,8$. The learnable parameters are initialized to zero, so the activation is initialized to be ReLU. This helps the training at the very beginning, exploiting all the nice properties of ReLU. For example, MeLU is convex for many iterations at the beginning of the training.

## 4. Experimental Results

We tested our novel activation function using the CNNs detailed in the previous section on a heterogeneous selection of publicly available datasets. In detail, the datasets are summarized in Table 2.
The protocol used in our experiments is a five-fold cross-validation, unless differently specified in the dataset description above. To validate the experiments the Wilcoxon signed rank test (Demsar, 2006) has been used.

Table 2. Descriptive Summary of the Datasets: the number of classes (#C), number of samples (#S)

| Dataset | Description | #C | #S | URL for Download |
|---|---|---|---|---|
| CH | The Chinese Hamster Ovary Cells dataset (Boland & Murphy, 2001) | 5 | 327 | https://ome.grc.nia.nih.gov/iicbu2008/hela/index.html |
| HE | the 2D HELA dataset (Boland & Murphy, 2001) | 10 | 862 | https://ome.grc.nia.nih.gov/iicbu2008/hela/index.html |
| LO | the Locate Endogenous dataset (Moccia et al., 2001) | 10 | 502 | |
| TR | the Locate Transfected dataset (Moccia et al., 2001) | 11 | 553 | |
| RN | the Fly Cell dataset (Shamir, Orlov, Eckley, Macura, & Goldberg, 2008) | 10 | 200 | https://ome.grc.nia.nih.gov/iicbu2008/hela/index.html |
| TB | Terminal bulb aging (Shamir et al., 2008) dataset of images of C. elegans terminal bulb at 7 ages | 7 | 970 | https://ome.grc.nia.nih.gov/iicbu2008 |
| LY | Lymphoma dataset (Shamir et al., 2008) | 3 | 375 | https://ome.grc.nia.nih.gov/iicbu2008 |
| MA | Muscle aging (Shamir et al., 2008). This dataset includes images of C. elegans muscles at 4 ages | 4 | 237 | https://ome.grc.nia.nih.gov/iicbu2008 |
| LG | Liver gender (Shamir et al., 2008). This dataset shows liver tissue sections from 6-month male and female mice on a caloric restriction diet, the 2 classes being male vs female. | 2 | 265 | https://ome.grc.nia.nih.gov/iicbu2008 |
| LA | Liver aging (Shamir et al., 2008). This dataset shows liver tissue sections from female mice on ad-libitum diet of 4 ages | 4 | 529 | https://ome.grc.nia.nih.gov/iicbu2008 |
| CO | histological images of human colorectal cancer (Kather et al., 2016). | 8 | 5000 | https://zenodo.org/record/53169#.WaXjW8hJaUm |
| BGR | breast grading carcinoma (Dimitropoulos et al., 2017) | 3 | 300 | https://zenodo.org/record/834910#.Wp1bQ-jOWUl |
| LAR | Laryngeal dataset (Moccia et al., 2017) | 3 | 1320 | https://zenodo.org/record/1003200#.WdeQcnBx0nQ |

In tables 4 and 5 we report the performance obtained using different activation functions coupled with Vgg16 and ResNet50. The performance is measured as the classification accuracy. To reduce the computation time all the results are calculates using a batch size (BS) of 30 and a learning rate (LR) of 0.0001 for 30 epochs. To improve the performance of the networks, we use data augmentation. Data augmentation is a technique that consists in changing the input image to create many new images that represent the same object. In this paper we randomly reflected the original images in both axis and rescaled them in both axis by two factors uniformly sampled in [1,2]. This means that the vertical and horizontal proportions of the new image are rescaled.

In Table 3 we summarize the initialization of the hyperparameters and of the learnable parameters of the activations that we use a baseline. We use the variable *maxInput* in the sense that we explained before to initialize the parameters in SReLU and to change the relative learning rate of the learnable parameters in APLU.

Table 3. Initialization of the hyperparameters

| Leaky ReLU | $a = 0.01$ |
|---|---|
| ELU | $a = 1$ |
| PReLU | $a_c = 0$ for every channel |
| SReLU | $a^l = 0, t^l = 0, a^r = 1, t^r = maxInput$, hence it is equal to ReLU in the first step |
| APLU | $a_c = 0$ for every channel, while $b_c$ are random |

Besides, the learning rate of the coefficients $a_c$ in APLU was divided by $maxInput$ to improve convergence. This means that, if λ is the global learning rate, the learning rate $λ^*$ of the parameters $a_c$ is given by

$$\lambda^* = \frac{\lambda}{maxInput}$$

We also created six ensembles of those networks by applying a sum rule. The sum rule consists in summing all the scores vectors of the networks in the ensemble to create a new score vector that takes into account the predictions of all the networks. The class predicted by the ensemble is the one with the highest score.

Our six ensembles are divided into:

- ENS: ensemble among all the methods with a given $maxInput$ of a given CNN. Hence, for a given CNN, there are two different ensembles named ENS, one for each choice of $maxInput$;
- eENS: ensemble among all the methods of a given CNN.

Table 4. Performance obtained using ResNet50.

| | Activation | CH | HE | LO | TR | RN | TB | LY | MA | LG | LA | CO | BG | LAR | Avg |
|---|---|---|---|---|---|---|---|---|---|---|---|---|---|---|---|
| ResNet50 MaxInput=1 | MeLU (k=8) | 92.92 | 86.40 | 91.80 | 82.91 | 25.50 | 56.29 | 67.47 | 76.25 | 91.00 | 82.48 | 94.82 | 89.67 | 88.79 | 78.94 |
| | Leaky ReLU | 89.23 | 87.09 | 92.80 | 84.18 | 34.00 | 57.11 | 70.93 | 79.17 | 93.67 | 82.48 | 95.66 | 90.33 | 87.27 | 80.30 |
| | ELU | 90.15 | 86.74 | 94.00 | 85.82 | 48.00 | 60.82 | 65.33 | 85.00 | 96.00 | 90.10 | 95.14 | 89.33 | **89.92** | 82.79 |
| | MeLU (k=4) | 91.08 | 85.35 | 92.80 | 84.91 | 27.50 | 55.36 | 68.53 | 77.08 | 90.00 | 79.43 | 95.34 | 89.33 | 87.20 | 78.76 |
| | PReLU | 92.00 | 85.35 | 91.40 | 81.64 | 33.50 | 57.11 | 68.80 | 76.25 | 88.33 | 82.10 | 95.68 | 88.67 | 89.55 | 79.26 |
| | SReLU | 91.38 | 85.58 | 92.60 | 83.27 | 30.00 | 55.88 | 69.33 | 75.00 | 88.00 | 82.10 | 95.66 | 89.00 | 89.47 | 79.02 |
| | APLU | 92.31 | 87.09 | 93.20 | 80.91 | 25.00 | 54.12 | 67.20 | 76.67 | 93.00 | 82.67 | 95.46 | 90.33 | 88.86 | 78.98 |
| | ReLU | 93.54 | 89.88 | 95.60 | 90.00 | 55.00 | 58.45 | **77.87** | 90.00 | 93.00 | 85.14 | 94.92 | 88.67 | 87.05 | 84.54 |
| | ENS | **95.38** | 89.53 | **97.00** | 89.82 | **59.00** | 62.78 | 76.53 | 86.67 | 96.00 | 91.43 | **96.60** | 91.00 | 89.92 | 86.28 |
| ResNet50 MaxInput=255 | MeLU (k=8) | 94.46 | 89.30 | 94.20 | 92.18 | 54.00 | 61.86 | 75.73 | 89.17 | 97.00 | 88.57 | 95.60 | 87.67 | 88.71 | 85.26 |
| | MeLU (k=4) | 92.92 | 90.23 | 95.00 | 91.82 | 57.00 | 59.79 | **78.40** | 87.50 | 97.33 | 85.14 | 95.72 | 89.33 | 88.26 | 85.26 |
| | SReLU | 92.31 | 89.42 | 93.00 | 90.73 | 56.50 | 59.69 | 73.33 | **91.67** | **98.33** | 88.95 | 95.52 | 89.67 | 87.88 | 85.15 |
| | APLU | 95.08 | 89.19 | 93.60 | 90.73 | 47.50 | 56.91 | 75.20 | 89.17 | 97.33 | 87.05 | 95.68 | 89.67 | 89.47 | 84.35 |
| | ReLU | 93.54 | 89.88 | 95.60 | 90.00 | 55.00 | 58.45 | 77.87 | 90.00 | 93.00 | 85.14 | 94.92 | 88.67 | 87.05 | 84.54 |
| | ENS | 93.85 | 91.28 | 96.20 | **93.27** | 59.00 | 63.30 | 77.60 | **91.67** | 98.00 | 87.43 | 96.30 | 89.00 | 89.17 | 86.62 |
| eENS | | 94.77 | **91.40** | **97.00** | 92.91 | **60.00** | **64.74** | 77.87 | 88.75 | 98.00 | 90.10 | 96.50 | 90.00 | 89.77 | **87.06** |

Table 5. Performance obtained using Vgg16.

| | Activation | CH | HE | LO | TR | RN | TB | LY | MA | LG | LA | CO | BG | LAR | Avg |
|---|---|---|---|---|---|---|---|---|---|---|---|---|---|---|---|
| Vgg16 MaxInput=1 | MeLU (k=8) | **99.69** | 92.09 | 98.00 | 92.91 | 59.00 | 60.93 | 78.67 | 87.92 | 86.67 | 93.14 | 95.20 | 89.67 | 90.53 | 86.49 |
| | Leaky ReLU | 99.08 | 91.98 | 98.00 | 93.45 | 66.50 | 61.13 | 80.00 | 92.08 | 86.67 | 91.81 | 95.62 | 91.33 | 88.94 | 87.43 |
| | ELU | 98.77 | **93.95** | 97.00 | 92.36 | 56.00 | 59.69 | 81.60 | 90.83 | 78.33 | 85.90 | 95.78 | 93.00 | 90.45 | 85.66 |
| | MeLU (k=4) | 99.38 | 91.16 | 97.60 | 92.73 | 64.50 | 62.37 | 81.07 | 89.58 | 86.00 | 89.71 | 95.82 | 89.67 | 93.18 | 87.13 |
| | PReLU | 99.08 | 90.47 | 97.80 | 94.55 | 64.00 | 60.00 | 81.33 | **92.92** | 78.33 | 91.05 | 95.80 | 92.67 | 90.38 | 86.79 |
| | SReLU | 99.08 | 91.16 | 97.00 | 93.64 | 65.50 | 60.62 | 82.67 | 90.00 | 79.33 | 93.33 | 96.10 | 94.00 | 92.58 | 87.30 |
| | APLU | 99.08 | 92.33 | 97.60 | 91.82 | 63.50 | 62.27 | 77.33 | 90.00 | 82.00 | 92.38 | 96.00 | 91.33 | 90.98 | 86.66 |
| | ReLu | **99.69** | 93.60 | 98.20 | 93.27 | **69.50** | 61.44 | 80.80 | 85.00 | 85.33 | 88.57 | 95.50 | 93.00 | 91.44 | 87.33 |
| | ENS | 99.38 | 93.84 | 98.40 | **95.64** | 68.00 | **65.67** | 85.07 | 92.08 | 85.00 | **96.38** | 96.74 | 94.33 | 92.65 | 89.47 |
| Vgg16 MaxInput=255 | MeLU (k=8) | **99.69** | 92.09 | 97.40 | 93.09 | 59.50 | 60.82 | 80.53 | 88.75 | 80.33 | 88.57 | 95.94 | 90.33 | 88.33 | 85.79 |
| | MeLU (k=4) | 99.38 | 91.98 | 98.60 | 92.55 | 66.50 | 59.59 | 84.53 | 91.67 | **88.00** | 94.86 | 95.46 | 93.00 | **93.03** | 88.39 |
| | SReLU | 98.77 | 93.14 | 97.00 | 92.18 | 65.00 | 62.47 | 77.60 | 89.58 | 76.00 | 96.00 | 95.84 | 94.33 | 89.85 | 86.75 |
| | APLU | 98.77 | 92.91 | 97.40 | 93.09 | 63.00 | 57.32 | 82.67 | 90.42 | 77.00 | 90.67 | 94.90 | 93.00 | 91.21 | 86.33 |
| | ReLu | **99.69** | 93.60 | 98.20 | 93.27 | **69.50** | 61.44 | 80.80 | 85.00 | 85.33 | 88.57 | 95.50 | 93.00 | 91.44 | 87.33 |
| | ENS | 99.38 | 93.84 | 98.80 | 95.27 | 68.50 | 64.23 | 84.53 | 92.50 | 81.33 | **96.57** | 96.66 | 95.00 | 92.20 | 89.13 |
| eENS | | 99.38 | **94.07** | 98.80 | 95.64 | 69.00 | 65.88 | 85.87 | 93.33 | 82.67 | **96.57** | 96.88 | 95.33 | 92.50 | **89.68** |

In Table 6 we report the results of the Wilcoxon signed-rank test on some of the proposed activations and ensembles. We did not consider those activations with $maxInput = 1$ in order to make the table smaller. One can see that the ensembles statistically differ from every single network with a p-value lower than 0.001.

Table 6. p-values of the classification task

| Activation | Leaky ReLU | ELU | PReLU | ReLU | MeLU (k=8) | MeLU (k=4) | SReLU | APLU | ENS | eENS |
|---|---|---|---|---|---|---|---|---|---|---|
| Leaky ReLU | ------- | 0.395 | 0.092 | 0.029 | 0.200 | 0.001 | 0.042 | 0.046 | 0.000 | 0.000 |
| ELU | ------- | ------- | 0.319 | 0.074 | 0.104 | 0.006 | 0.095 | 0.123 | 0.000 | 0.000 |
| PReLU | ------- | ------- | ------- | 0.009 | 0.096 | 0.001 | 0.025 | 0.033 | 0.000 | 0.000 |
| ReLU | ------- | ------- | ------- | ------- | 0.568 | 0.103 | 0.829 | 0.809 | 0.000 | 0.000 |
| MeLU (k=8) | ------- | ------- | ------- | ------- | ------- | 0.082 | 0.611 | 0.726 | 0.000 | 0.000 |
| MeLU (k=4) | ------- | ------- | ------- | ------- | ------- | ------- | 0.213 | 0.059 | 0.001 | 0.000 |
| SReLU | ------- | ------- | ------- | ------- | ------- | ------- | ------- | 0.616 | 0.000 | 0.000 |
| APLU | ------- | ------- | ------- | ------- | ------- | ------- | ------- | ------- | 0.001 | 0.000 |
| ENS | ------- | ------- | ------- | ------- | ------- | ------- | ------- | ------- | ------- | 0.002 |
| eENS | ------- | ------- | ------- | ------- | ------- | ------- | ------- | ------- | ------- | ------- |

In Table 7 we report performance, in some datasets, obtained choosing optimal values of BS and LR for ReLU. Also with BS and LR optimized for ReLU the performance of ENS is higher than that obtained by ReLU. This means that in this experiment we somehow overfit the networks trained with ReLU choosing those BS and LR that obtained the best performance in the test set. We show that also in this case the ensemble outperform stand-alone ReLU.

Table 7. Performance with optimized BS and LR.

| | Activation | CH | LA | | MA |
|---|---|---|---|---|---|
| Resnet50 MaxInput=255 BS=10 LR=0.001 | MeLU (k=8) | 98.15 | 98.48 | Vgg16 MaxInput=255 BS=50 LR=0.0001 | 90.42 |
| | MeLU (k=4) | 98.15 | 98.67 | | 87.08 |
| | SReLU | 99.08 | 96.00 | | 88.33 |
| | APLU | 98.46 | 98.48 | | 93.75 |
| | ReLu | 97.23 | 96.57 | | 92.08 |
| | ENS | **99.38** | **99.05** | | **93.75** |

Besides, for a more fair comparison, we tested an ensemble created with different activations and an ensemble created using ReLU networks. We followed the approach proposed in (Lumini, Nanni, & Maguolo, 2020) and created a learned ensemble of networks whose output vector is the weighted sum of the output vectors of the networks in the ensemble. The method consists in minimizing the average crossentropy loss among all the test sets that we considered, except from one left out test set. Then we test the ensemble on the left out test set. Besides, with add a concave penalty to the weights $w_i$ such that it encourages the learning of ensembles whose weights are zero, i.e: are smaller in size. This penalty is equal to

$$L^{REG} = \sum w_i^\gamma$$

where $\gamma < 1$. By changing the value of $\gamma$ we were able to create ensembles of different sizes ranging from 6 to 19 among a pool of 24 networks. We repeated this experiment using only 24 ReLU networks in the first scenario (RELU ENS) and the 24 networks of Tables 4 and 5 in the second one (ACT ENS). In Figure 2 we show the average performances of the two methods for different ensemble sizes. One can see that our approach outperforms the ReLU baseline every time. Apparently, the best ensembles never had a size larger than 19, probably due to the fact that some networks had a low performance. This is consistent with the fact that eENS contains all the networks and it performs worse than the new approaches. We want to stress the fact that the trainings of the ensemble are independent from each other, i.e: the ensemble of size $n$ is not learned starting from the ensemble of size $n - 1$.

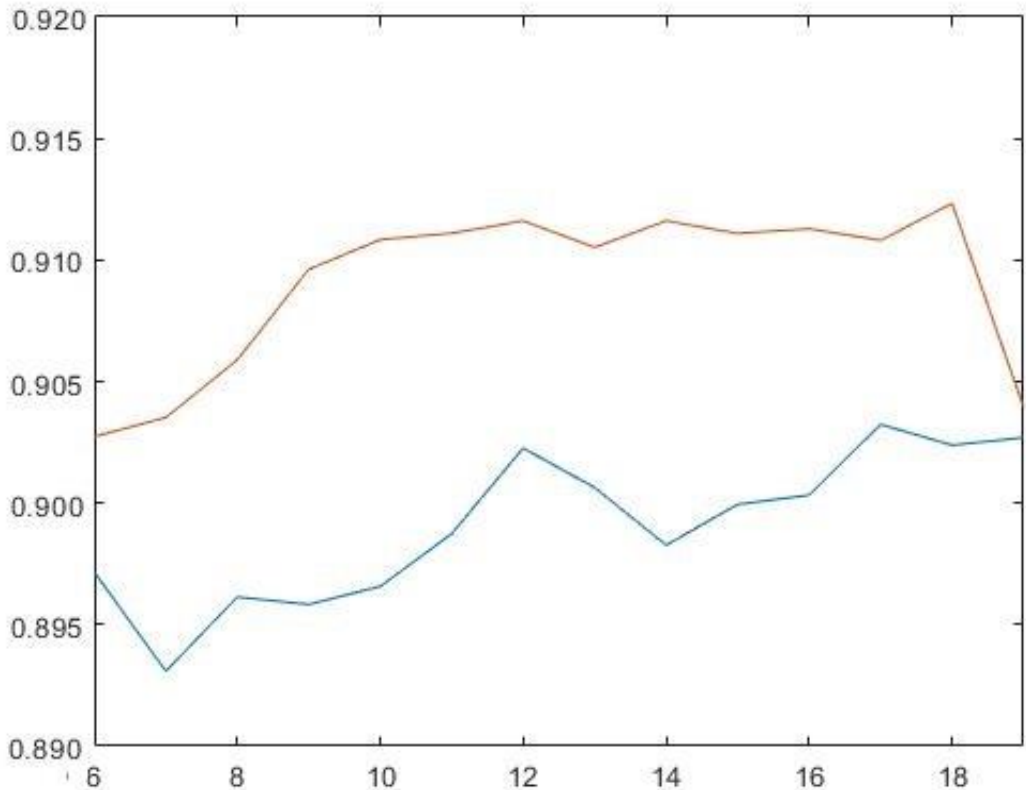

Figure 2. Performances of RELU ENS (blue line) and ACT ENS (red line). The size of the ensemble is on the horizontal axis and the relative performance on the vertical axis.

Every additional parameter in the activation functions increases the computational time for training and testing. We report the training and test times for one fold of two datasets (CH, HE) in Table 8. Times are reported in seconds. In both datasets we trained and tested VGG16 on one fold.
ReLU is by far the fastest, although it is an unfair comparison, because it is the only function completely created by MATLAB developers. Our implementations of the other functions might be not optimal. One can see that MeLU with $k = 8$ is the slowest function. However, the speed of MeLU with $k = 4$ is comparable to the one of the other activation functions.

Table 8. Training and test times for the different activations using VGG16.

|    |       | ReLU | Leaky ReLU | ELU | PReLU | MeLU (k=8) | MeLU (k=4) | SReLU | APLU |
|----|-------|------|------------|-----|-------|------------|------------|-------|------|
| CH | Train | 81   | 130        | 137 | 138   | 463        | 288        | 361   | 225  |
|    | Test  | 0.34 | 0.6        | 0.9 | 0.7   | 2.6        | 1.5        | 1.6   | 1.3  |
| HE | Train | 224  | 340        | 352 | 365   | 1230       | 777        | 993   | 612  |
|    | Test  | 0.58 | 0.89       | 1.2 | 1.0   | 3.1        | 2.12       | 2.75  | 1.6  |

In general, adding parameters make the network slower, but not by many order of magnitude. If training and inference times are not particularly relevant, learnable activations can be useful to improve the network performance.

From the results reported in Tables 4, 5, 6 & 7 the following conclusions can be drawn:
- both ENS and eENS outperform with a p-value lower than 0.02 all the stand-alone activation functions. Moreover, eENS outperforms ENS in both the CNN topologies (i.e. Vgg16 and ResNet50) with a p-value of 0.05. This is the most important finding of this work;
- MeLU obtains the best average p6terformance in both the CNNs;
- Different behaviors occur in the two topologies, since in ResNet50 there is a clear performance difference between $MaxInput = 1, 255$, while in Vgg16 similar performance is obtained with $MaxInput = 1, 255$.
- Also optimizing BS and LR for ReLU similar conclusions are obtained, ENS outperforms the other activation functions, including ReLU;
- From our experiments one can see that the best approaches depend on the applications. If the training time is limited, one should use ReLU; if speed at inference time is important, we suggest to use MeLU. Otherwise, if only the accuracy matters, we suggest to create an ensemble.

**Conclusion**

The purpose of the present paper was to evaluate the performance of an ensemble of CNNs created by changing the activation functions in famous pre-trained networks. Besides, we tested several activation functions on several challenging datasets and reported their results. We also proposed a new activation function called Mexican Linear Unit.

Our experiments show that an ensemble of multiple CNNs that only differ in the activation functions outperforms the results of the single CNNs and of naïve ensembles made by ReLU networks. Besides, we show that there is not an activation that is consistently better than the others. In particular, we see that MeLU is competitive with the other activation functions in the literature. MeLU also seems to be the best performing activation when $k = 4$, in particular on VGG16. Notice that we only tested MeLU with $k = 4,8$, we did not cherry-picked the best performing parameters $k$ on the test set. As future work we aim to create even larger ensembles of CNNs to see how much we can boost the performances of the single CNN. The drawbacks of this approach are speed and memory requirements. However, we plan to do it with very small CNNs and see if such an ensemble is competitive with much larger networks which are still larger than the ensemble.

Finally, we share the MATLAB code of every activation and ensemble that we created.

**Acknowledgment**

We gratefully acknowledge the support of NVIDIA Corporation for the "NVIDIA Hardware Donation Grant" of a Titan X used in this research.

Recognition. *CoRR*, *abs/1409.1*.